\pdfoutput=1

\documentclass[11pt]{article}

\usepackage{times}
\usepackage[final]{acl}
\usepackage{latexsym}
\usepackage{array}

\usepackage[T1]{fontenc}

\usepackage[utf8]{inputenc}

\usepackage{microtype}

\usepackage{inconsolata}

\usepackage{graphicx}
\usepackage{booktabs}
\usepackage{multirow}
\usepackage{amsmath}    
\usepackage{amssymb}    
\usepackage{float}      
\usepackage{bm}         
\title{{DiaTool-DPO}: Multi-{Turn} {Direct} {Preference} {Optimization} for {Tool-Augmented} {Large} {Language} {Models}}

\author{Sunghee Jung, Donghun Lee, Shinbok Lee, \\
{\bf Gaeun Seo, Daniel Lee, Byeongil Ko,}\\ 
{\bf  Junrae Cho, Kihyun Kim, Eunggyun Kim, and Myeongcheol Shin} \\
Kakao Corp. \\
Seongnam-si, Gyeonggi-do, South Korea\\
\texttt{kong.2024@kakaocorp.com} \\
}

\begin{document}
\maketitle
\begin{abstract}
Tool-Augmented Larage Language Models (TA-LLMs) have shown promise in real-world applications, but face challenges in handling incomplete queries and out-of-scope requests. While existing approaches rely mainly on Supervised Fine-Tuning with expert trajectories, we propose DiaTool-DPO, a novel method that enhances TA-LLM's dialogue capabilities through Direct Preference Optimization. We model TA-LLM interactions as a Markov Decision Process with 5 distinct dialogue states and categorize user queries into 3 types based on their state transition trajectories. We automatically construct paired trajectory datasets of correct and incorrect dialogue flows and introduce a specialized objective loss for dialogue control. Our comprehensive evaluation demonstrates that DiaTool-DPO approaches GPT-4o's performance (94.8\% in information gathering, 91\% in tool call rejection) with substantial improvements over baseline (44\% and 9.6\% respectively) while maintaining core functionality. Our approach opens new possibilities for developing TA-LLMs that can handle diverse real-world scenarios without requiring additional expert demonstrations or human labeling.
\end{abstract}
\section{Introduction}
The conversational capabilities of Tool-Augmented Large Language Models (TA-LLMs) are crucial. Specifically, they must be able to control conversation flow by determining whether to 1) ask follow-up questions for clarification from the user or 2) make a tool call or 3) reject tool calls when no suitable tools(functions, used interchangeably) are available. In particular, failing to generate follow-up questions or reject tool calls leads to the risk of invoking tool calls with hallucinated information \citep{huang-etal-2024-metatool, yang2024can}.  While early benchmarks focused on successful tool calls through proper tool selection and argument extraction \citep{t-eval2024,ye2024,farn2023,bfcl2024}, recent benchmarks increasingly evaluate the ability to engage in user conversations \citep{huang-etal-2024-metatool,yang2024can,kakao2024functionchat}. However, we find that Supervised Fine-Tuning (SFT) techniques alone has its limitation to learn conversational skills\citep{yang2024can,qin2024toolllm,patil2024gorilla,tang2023toolalpaca}. 
While research applying techniques beyond SFT for training TA-LLMs is limited, many other areas of LLM research successfully utilize reinforcement learning to capture subtle human preferences \citep{kaufmann2023survey,nguyen2024reinforcement}. Notably, many works solved agent tasks such as WebShop, ALFWorld, and ScienceWorld with reinforcement learning techniques due to their explicitly defined rewards in the datasets. \citep{yao2022webshop,shridhar2021alfworld,wang-etal-2022-scienceworld,clarify2024,toolfeedback2024,trialerror2024,dmpo2024}

Here, we formulate TA-LLM task as a Markov Decision Process(MDP) and adapt reinforcement learning techniques to enhance TA-LLMs' conversational abilities. We hypothesize that TA-LLMs have five internal states and classify user queries into three types based on state transition trajectories \citep{bellman1957markovian,howard1960dynamic,puterman1994markov,sutton1998reinforcement}. 
We automatically generate rejected trajectories by pairing user queries with mismatched conversation trajectories.
The resulting dataset $D$ consists of paired trajectories ($\tau_c, \tau_r$), where $\tau_c$ represents the chosen (preferred) trajectory and $\tau_r$ denotes the rejected (non-preferred) trajectory for each training instance. While our primary experiments were conducted in Korean, we demonstrate that our approach is language-agnostic and can be applied to English as well.
This dataset is termed the \textbf{DiaTool-DPO (Dialogue Tool DPO)} dataset, with its corresponding training algorithm named DiaTool-DPO. To comprehensively evaluate the impact of this dataset and algorithm, we assess not only slot-filling and tool call rejection capabilities but also all other general competencies across various base LLM models.\\
In this paper, we leverage Direct Preference Optimzation(DPO) techniques to improve conversational abilities by controlling TA-LLM's dialogue flow \citep{rafailov2023direct_}. Specifically, the contributions of this paper are as follows:
\begin{itemize}
\item \textbf{Formulation as a Reinforcement Learning Problem: } 
We suggested a novel approach that formulates tool-learning via reinforcement learning by defining hidden internal states and query classes, and solves the resulting Markov decision process through DPO.
\item \textbf{Dataset Construction:} We automatically construct the DiaTool-DPO (Dialogue Tool DPO) dataset that enables TA-LLMs to learn which conversation flow to choose in specific situations without human labor.
\item \textbf{Novel Alignment Objective:} We propose a specialized alignment objective for TA-LLMs that controls conversation flow through the contrast between chosen and rejected trajectories.
\end{itemize}
The source code of this work will be released.\footnote{https://github.com/kakao/diatool-dpo}
\section{Related Works}
\label{sec:related_works}
\paragraph{Benchmarks for TA-LLMs}
Early benchmarks (T-Eval \citep{t-eval2024}, ToolEyes \citep{ye2024}, BFCL \citep{bfcl2024}) focused on tool call evaluation. Recent ones emphasize conversational abilities, including tool awareness \citep{huang-etal-2024-metatool}, query disambiguation \citep{yang2024can}, and multi-turn dialogue (BFCL v3 \citep{bfcl2024}, API-Bank \citep{li-etal-2023-api}, ToolSandbox \citep{lu2024toolsandbox}). FunctionChat-Bench \citep{kakao2024functionchat} provides a comprehensive evaluation framework for tool-related capabilities.

\paragraph{Reinforcement Learning for Agent Tasks}
Recent work has explored reinforcement learning for improving LLM-based agents. \citet{clarify2024} addressed disambiguation using quasi-online DPO with user simulation \cite{rafailov2023direct}. \citet{toolfeedback2024} enhanced tool usage through ranking-based learning that considers response-ground truth distances. \citet{trialerror2024} combined SFT with offline exploration, creating DPO datasets from expert and failure trajectories. \citet{dmpo2024} and \citet{xiong2025building} introduced multi-turn DPO-based learning methods for language agents and math agents, respectively.
\begin{figure*}[t]
    \centering
    \includegraphics[width=\textwidth]{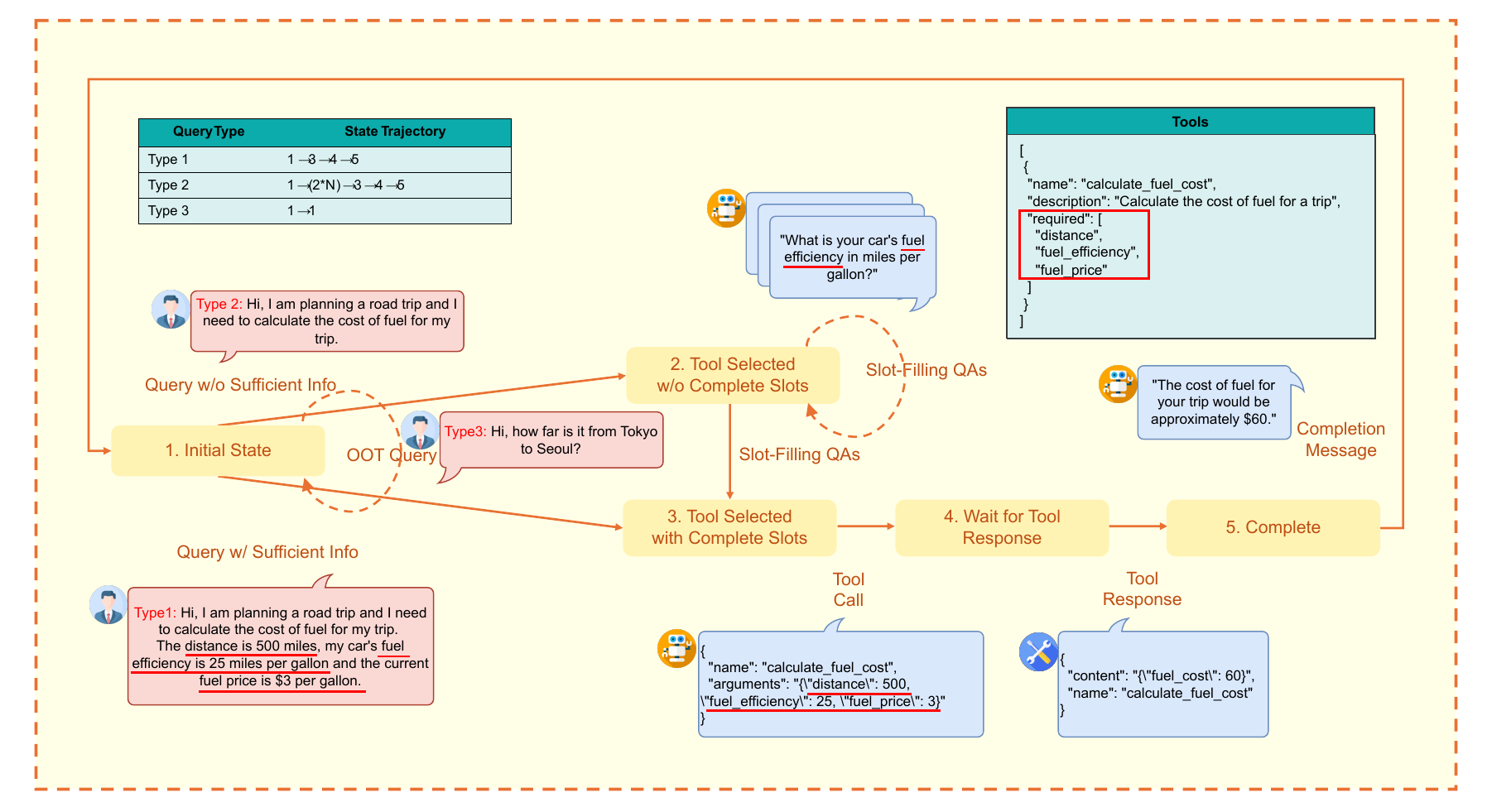}
    \caption{Visualization of five internal states of TA-LLMs and state trajectories for three different query types. User queries are shown in red message bubbles, while other conversational turns are displayed in blue. OOT (Out-of-Tools) queries represent requests for functionality not available in the teal-colored "Tools" list. Slot-Filling QAs denote conversational turns aimed at gathering required fields for tool execution. Tool calls represent messages where the assistant invokes a tool, tool responses show the returned execution results, and completion messages demonstrate the assistant's final response using the tool output. For optimal visualization of the state transitions and message types, we recommend viewing this figure in color. Icons from Flaticon\protect\footnotemark are used in this diagram.}
    \label{state_definition}
\end{figure*}
\footnotetext{http://www.flaticon.com}

\section{Preliminaries}
\begin{table*}[t]
\footnotesize
\centering
\begin{tabular}{@{}llllrl@{}}
\toprule
\textbf{Query} & \textbf{Chosen traj.} & \textbf{Rejected traj.} & \textbf{Learning lesson} & \textbf{Count} & \textbf{Included} \\
\midrule
\multirow{4}{*}{Type 1} & \multirow{4}{*}{1→3→4→5} & 1→2→3→4→5 & Prevent redundant slot-filling & 2,089 & Easy \\
& & 1→(2*N)→3→4→5 & Prevent redundant slot-filling & 562 & Hard \\
& & 1→(2*M)→3→4→5 & Prevent redundant slot-filling & 2,530 & Hard \\
& & 1→1 & Tool call accept & 2,090/562 & Easy, Hard \\
\midrule
\multirow{4}{*}{Type 2} & 1→2→3→4→5 & 1→3→4→5 & Prevent slot hallucination & 2,089 & Easy \\
& 1→(2*N)→3→4→5 & 1→3→4→5 & Prevent slot hallucination & 562 & Hard \\
& & 1→(2*M)→3→4→5 & Prevent slot hallucination & 2,530 & Hard \\
& & 1→1 & Tool call accept & 2,089/562 & Easy, Hard \\
\midrule
\multirow{2}{*}{Type 3} & \multirow{2}{*}{1→1} & 1→3→4 & Tool call reject & 567 & Hard \\
& & 1→(2*N)→3→4 & Tool call reject & 562 & Hard \\
\bottomrule
\end{tabular}
\caption{Training data composition and corresponding learning objectives. Each query type is defined in Section \ref{subsubsec:Definition_of_TA-LLM_query_types}. In the trajectory notation, $(2*N)$ denotes that state 2 is visited $N$ consecutive times (e.g., $2\to2\to2$ for $N=3$), where $N$ represents the number of total unknown required fields. $M$ denotes the size of a subset of unknown required fields ($N > M > 1$). In the Count column, when both difficulty levels are included, values are presented as "number of Easy examples/number of Hard examples"}
\label{tab:training_data}
\end{table*}
\subsection{Definition of TA-LLM Internal States}
As shown in Figure \ref{state_definition}, we define five inner states for TA-LLMs. These states are not observed externally. We arbitrarily hypothesized states with the objective of contrasting chosen trajectory paths and rejected trajectory paths for slot-filling and tool call rejection problems.
The traversed states and their trajectories vary depending on the query type.
The system falls into one of the following states in its tool execution lifecycle.
\paragraph{Initial State }A state with no history. If the user's request cannot be supported by any tool in the available tool list, the system returns a tool call rejection message and remains in this state.
\paragraph{Tool Selected without Complete Slots }A state where the dialogue history has provided information about which tool to select, but not all required parameters for tool invocation. Through slot-filling QA interactions, the system can transition to the next stage. If multiple required fields need to be determined, this stage may be repeated.
\paragraph{Tool Selected with Complete Slots }A state where the dialogue history has provided both the tool selection and all necessary argument values. The system can transition to this state either directly from the \textit{Initial State} or through the \textit{Tool Selected without Complete Slots} state.
\paragraph{Wait for Tool Response} A state where the tool call has been executed and the system is awaiting execution results from the tool.
\paragraph{Complete }A state reached after receiving the tool's execution results, prior to generating a \textit{completion message} containing the tool execution outcome for the user.
\subsection{Classification of TA-LLM Query Types}
\label{subsubsec:Definition_of_TA-LLM_query_types}
We classified queries into three types based on conversation trajectories:
\begin{itemize}
\item \textbf{Type 1:} Queries that contain all necessary information for tool calls in the initial user query, enabling immediate tool calls without slot-filling.
\item \textbf{Type 2:} Queries that lack argument information for tool calls in the initial user query, requiring slot-filling before making tool calls.
\item \textbf{Type 3:} Queries where the requested functionality is not available in the TA-LLM's capability list (referring to the \textit{tools} shown in Figure \ref{state_definition}), necessitating rejection of the tool call.
\end{itemize}
In Figure \ref{state_definition}, the "calculate\_fuel\_cost" function in "Tools" has required fields: distance, fuel efficiency, and fuel price. Type 1 queries already contain these three pieces of information, eliminating the need for slot-filling questions. Consequently, they skip State 2 and proceed directly to State 3. In State 3, with all necessary information prepared for the tool call, the call is executed, and the system advances to State 4. State 4 waits for the tool response, and upon receiving it, transitions to State 5. In State 5, the system returns a completion message before returning to the State 1.
Type 2 queries lack all three required fields and thus remain in State 2 while slot-filling occurs until all required fields are known, at which point they transition to State 3. Subsequent transitions follow the same path as Type 1 queries.
Type 3 queries are unrelated to any tool in the available toolset, resulting in a tool call rejection message, with the TA-LLM remaining in State 1.
\subsection{Evaluation}
Our primary focus in this paper is the conversational ability of TA-LLMs, specifically concerning slot-filling and tool-rejection aspects. However, we also evaluate other metrics to ensure that enhancing these abilities does not compromise other fundamental performance indicators. We adopted the open-source benchmark FunctionChat-Bench \citep{kakao2024functionchat}. This is because, to the best of our knowledge, this benchmark is the only publicly available open-source benchmark that includes slot-filling ability. It evaluates accuracies for following aspects of tool calls.
\begin{itemize}
\item \textbf{Call:} This metric evaluates whether the correct tool was selected and called with accurate arguments
\item \textbf{Completion:} This metric assesses the ability to convert tool responses into appropriate text-based answers, termed \textit{completion messages}, that address the user's initial query
\item \textbf{Slot:} This metric evaluates whether appropriate questions were asked when the user query lacked necessary arguments for tool calls
\item \textbf{Relevance:} This metric assesses whether the system can properly decline requests when the required tool for answering the user query is not available in the \textit{tools} list
\end{itemize}

\section{Dataset Construction}
\subsection{Seed Trajectory Construction}
We will create the DiaTool-DPO dataset by pairing chosen trajectories with rejected trajectories as shown in Table \ref{tab:training_data}. Before this, we first generate a seed trajectory dataset (chosen trajectory dataset) that matches query Types 1, 2, and 3.
Using the glaive-function-calling-v2 dataset \citep{glaive2023function}, commonly known as glaive 2.0, we created the seed trajectory dataset. While each data point in the glaive 2.0 dataset corresponds to either Type 1 or Type 3, we need to construct matching chosen and rejected pairs as in Table \ref{tab:training_data}. Therefore, we sample Type 1 dialogues and augment them to create corresponding Type 2 and Type 3 trajectories. 
We first sampled a portion of Type 1 queries from the glaive 2.0 based on difficulty levels. These queries were then modified into incomplete queries requiring slot-filling through ChatGPT \citep{openai2023chatgpt} prompting, and augmented with subsequent slot-filling conversations to create matching Type 2 trajectories (See Appendix~\ref{sec:slot-fill-prompt} for the detailed prompt). We then generated corresponding Type 3 data by removing the appropriate tool from the tools list in both Type 1 and Type 2 samples.
We constructed the DiaTool-DPO dataset by matching pairs from these Type 1, Type 2, and Type 3 trajectory triplets. The dataset composition follows three distinct approaches based on difficulty levels, which we detail in Section \ref{subsubsec:Stratification by Difficulty}.
\subsection{Dataset Structure}
In the original DPO, single-turn problems are addressed. The dataset is structured as triplets of \{prompt, chosen, rejected\}. Here, prompt refers to the shared user query, while chosen and rejected represent desirable and undesirable answers, respectively.
In contrast, DiaTool-DPO deals with multi-turn problems. In the multi-turn trajectories addressed by DiaTool-DPO, aside from the initial user query, the remaining user queries are not shared between chosen and rejected trajectories. Accordingly, we structured the DiaTool-DPO dataset as pairs of \{chosen trajectory, rejected trajectory\}. Each trajectory features alternating user turns and assistant turns. The chosen trajectory and rejected trajectory have the same user query only in the first utterance. During training, user turns from each trajectory are not included in the loss calculation.

\subsection{DiaTool-DPO Dataset Composition by Difficulty Levels}
\label{subsubsec:Stratification by Difficulty}
We stratified our dataset by difficulty, creating two subsets: \textit{Easy} and \textit{Hard}. The aggregation of these subsets constitutes our complete dataset, denoted as \textit{All} throughout this paper. All experimental evaluations were conducted using the \textit{All} dataset unless explicitly specified otherwise. Specifically, Easy comprises 8,357 samples and Hard contains 8,437 samples, collectively forming the All dataset with a total of 16,794 instances.

\paragraph{Easy}
In the Easy dataset, slot-filling questions occur at most once per dialogue. As shown in Table \ref{tab:turn_counts}, chosen trajectories with slot-filling average 3 turns, while rejected trajectories average 2 turns, consisting only of a tool call and tool completion without the necessary slot-filling QA.  
Notably, as shown in Tables \ref{tab:training_data} and \ref{tab:turn_counts}, the Easy dataset excludes Type 3. However, as demonstrated in Type 1 and Type 2 entries of Table \ref{tab:training_data}, we incorporated \textit{1→1} as rejected trajectories, enabling indirect learning about tool call rejection through tool call accept
\paragraph{Hard}
The Hard dataset extends the Easy dataset by introducing complex slot-filling scenarios and tool call rejection cases. It features Type 1 trajectories with three or more required fields and Type 2 trajectories where users provide partial information incrementally. For Type 3, we created scenarios requiring tool call rejection by modifying the available tool list. In Table \ref{tab:turn_counts}, while chosen trajectories in relevance data average 1.0 turns due to immediate rejection, rejected trajectories average 2.3 turns.\\ 
In summary, the Hard dataset requires more sophisticated handling of slot-filling interactions and demonstrates higher average turn counts compared to the Easy dataset.

\begin{table}[htbp]
\footnotesize
\centering
\begin{tabular}{@{}llcc@{}}
\toprule
\textbf{Difficulty} & \textbf{Trajectory} & \textbf{Slot} & \textbf{Relevance} \\
\midrule
\multirow{2}{*}{Easy} & Chosen & 3.05 & N/A \\
& Rejected & 2.00 & N/A \\
\midrule
\multirow{2}{*}{Hard} & Chosen & 4.83 & 1.00 \\
& Rejected & 3.17 & 2.33 \\
\midrule
\multirow{2}{*}{All} & Chosen & 4.11 & 1.00 \\
& Rejected & 2.80 & 2.33 \\
\bottomrule
\end{tabular}
\caption{Average number of turns for chosen and rejected trajectories across different difficulty levels. Note that relevance metrics are only applicable for Hard and All difficulty samples.}
\label{tab:turn_counts}
\end{table}

\section{Training Objective}

Equation \ref{eq} presents the proposed objective loss for learning DiaTool-DPO.
$\pi_{ref}$ is a reference model obtained from SFT and $\pi_{\theta}$ is the model under training with $L_{align}$.
The chosen trajectory is defined as $\tau^c=\{s_0^c,a_0^c,s_1^c,a_1^c,...,s_{T_c}^c,a_{T_c}^c\}$, and the rejected trajectory is represented as $\tau^r=\{s_0^r,a_0^r,s_1^r,a_1^r,...,s_{T_r}^r,a_{T_r}^r\}$.
$s_t^c, s_t^r$ represent the user's utterances at dialogue turn $t$ in the chosen or rejected trajectory. This includes the \textit{tool response} from executing a call. $a_t^c, a_t^r$ represent the TA-LLM's responses at dialogue turn $t$ in the chosen or rejected trajectory.
 $s_0^c==s_0^r$ contains 1) the user's initial request corresponding to Type 1, 2, or 3, and 2) the list of tools available to the TA-LLM. $s_t^c,\ s_t^r, \ where \ t\neq0$ contain either 1) user responses to slot-filling questions or 2) tool responses. $a_t^c,\ a_t^r$ is one of: a slot-filling question, a tool call, or a completion message.
$T_c, T_r$ refer to the turn length in chosen trajectory and rejected trajectory, respectively. The turn weight $\phi$, the turn length normalization factor $\psi$, and the reward gap margin $\rho$ are explained as following. 

\begin{equation}
\begin{aligned}
L_{align} = &-\mathbb{E}_{(s_0,\tau^c,\tau^r)\sim D}\log \sigma\Big[ \\
    &\sum_{t=0}^{T_c-1} \beta\frac{\phi(t,T_c)}{\psi(T_c)}\log \frac{\pi_\theta(a_t^c|s_t^c)}{\pi_{ref}(a_t^c|s_t^c)} \\
    &- \sum_{t=0}^{T_r-1} \beta\frac{\phi(t,T_r)}{\psi(T_r)}\log \frac{\pi_\theta(a_t^r|s_t^r)}{\pi_{ref}(a_t^r|s_t^r)}-\rho\Big] \\
\text{where,}\\&\psi(T) = \sum_{t=0}^{T-1} \frac{1-\gamma^{T-t}}{1-\gamma^T},\\
&\phi(t,T) = \frac{1-\gamma^{T-t}}{1-\gamma^T}\\
&\rho = \text{arbitrary margin}
\end{aligned}
\label{eq}
\end{equation}

\paragraph{Turn Weight $\phi$}
To compare chosen and rejected trajectory of different turn length, we need to aggregate each trajectory into a single score. In this process, we assume that the first turn is more important than the last turn since the erroneous answer in first turn must lead to wrong answers in following conversations. Therefore, we introduced a turn weight $\phi$
, which decreases by $\gamma$ as turn $t$ increases and is designed to sum to 1. In a related study, $\phi$ was designed to cancel out partition function in DPO and their $\phi$ met all our requirements \cite{dmpo2024}. Thus, we adopted the weight $\phi$ 
as proposed in their research.
\paragraph{Turn Length Normalization $\psi$}
For slot-filling, rejected trajectories always have fewer turns due to partial or complete omission of slot-filling QAs, making chosen trajectories consistently longer. Conversely, for relevance, chosen trajectories complete with a tool call rejection in one turn, while rejected trajectories continue conversations, making rejected trajectories consistently longer.
As a result of turn length imbalance, there's an inherent bias towards larger chosen rewards when training for slot-filling. Likewise, there's a bias towards larger rejected rewards when training for relevance. Such weight bias based on turn length is undesirable. 
Based on this observation, we propose $\psi$, the sum of weights attributed to turn length, as a normalization term.\\
\paragraph{Reward Gap Margin $\rho$}
To control the difficulty of the learning problem presented to the model, we subtracted an additional margin from the reward gap between chosen and rejected trajectories. The effectiveness of subtracting a margin has been previously demonstrated in the SimPO \citep{Meng2024SimPO}.
\begin{table*}[t]
\footnotesize   
\centering
\begin{tabular}{lcccccc}
\toprule
\textbf{Description} & \textbf{Call} & \textbf{Completion} & \textbf{Slot} & \textbf{Relevance} & \textbf{Micro Avg.} & \textbf{Macro Avg.} \\
\midrule
SFT-only                                    & 0.843 & \textbf{0.957} & 0.639 & 0.826 & 0.844 & 0.816 \\
SFT w/ preferred responses                  & 0.457 & 0.900 & 0.806 & 0.913 & 0.725 & 0.769 \\
DiaTool-DPO w/o \{$\phi$, $\psi$, $\rho$\}  & 0.857 & 0.943 & 0.806 & 0.870 & 0.879 & 0.869 \\
DiaTool-DPO w/o \{$\psi$\,$\rho$\}          & 0.857 & 0.943 & 0.778 & 0.870 & 0.874 & 0.862 \\
DiaTool-DPO w/o $\rho$                      & 0.843 & 0.929 & 0.833 & 0.870 & 0.874 & 0.869 \\
DiaTool-DPO w/o $\psi$                      & \textbf{0.886} & \textbf{0.957} & 0.778 & \textbf{0.913} & 0.894 & 0.883 \\
DiaTool-DPO (Ours)                          & 0.857 & 0.929 & \textbf{0.917} & \textbf{0.913} & \textbf{0.905} & \textbf{0.904} \\
\midrule
GPT-4o-mini-2024-07-18                      & \textbf{0.929} & \textbf{0.971} & \textbf{0.972} & 0.739 & 0.920 & 0.903 \\
GPT-4o-2024-08-06                           & 0.914 & 0.926 & \textbf{0.972} & \textbf{1.000} & \textbf{0.925} & \textbf{0.953} \\
\bottomrule
\end{tabular}
\caption{Ablation study results of DiaTool-DPO comparing different model variants. ``SFT-only'' represents the model before DPO training, and ``SFT w/ preferred responses'' indicates training with only chosen responses from the DPO dataset. We systematically remove key components ($\phi$: reward scaling, $\psi$: total turn-length normalization, $\rho$: reward gap threshold) from our full model to analyze their individual contributions. GPT-4 models are included as reference points.}
\label{table:ablation}
\end{table*}
\section{Experiments}
\subsection{Experimenetal Setup}
\label{subsec:Experimental_setup}
Unless otherwise specified, we used LLaMA3-8B-Instruct as the baseline model \citep{meta2024llama3}. All models undergo two sequential training phases prior to DiaTool-DPO: continual pretraining(CPT) and SFT.
Details of CPT, SFT and training setup are available at Appendix~\ref{sec:SFT} and \ref{sec:training_setup}.
The DiaTool-DPO was trained on 95\% of the dataset, with the remaining 5\% reserved for validation. To rigorously evaluate the model's generalization performance, we conducted testing on the independent FunctionChat-Bench dataset, which remained completely isolated from both training and validation processes \citep{kakao2024functionchat}.
\subsection{Ablation Study}
\label{subsec:ablation_study}
Table \ref{table:ablation} demonstrates the impact of each DiaTool-DPO component on various performance metrics. All scores are normalized to a maximum of 1.0.
Since DiaTool-DPO datasets contain expert trajectories in the form of chosen trajectories, we needed to validate whether performance improvements stem from the DiaTool-DPO itself rather than mere increased exposure to expert trajectories. To this end, we conducted an "SFT w/ preferred responses" experiment, which applies further SFT to the "SFT-only" model using chosen trajectories extracted from the DiaTool-DPO dataset. Results show improvements in slot and relevance scores but a significant 45\% drop in call performance. We hypothesize that while SFT exposes the model to glaive 2.0-augmented expert trajectories with slot-filling QA, it fails to contrastively learn contextual differences between performing slot-filling QAs and making tool calls.

"DiaTool-DPO w/o \{$\phi$, $\psi$ and $\rho$\}" showed similar results to "SFT w/ preferred responses" but notably avoided the drops in call and completion metrics. This preservation of performance can be attributed to \{Type 1 - Type 2\} and \{Type 1 - Type 3\} trajectories in Table \ref{tab:training_data}, preventing inappropriate slot-filling or LLM dialogue in place of call situations. The comparison between these two approaches demonstrates that DPO-based methodology effectively learns context-dependent dialogue flow control.

Subsequent experiments examined the contribution of individual components within the DiaTool-DPO. Contrary to the findings of DMPO \citep{dmpo2024}, comparing "DiaTool-DPO w/o \{$\phi$, $\psi$ and $\rho$\}" and "DiaTool-DPO w/o \{$\psi$,$\rho$\}" revealed minimal impact of reward scaling on evaluation metrics. We attribute this discrepancy to different evaluation approaches between agent tasks and TA-LLM tasks : Agent tasks accumulate errors through trajectories, making initial turns more important than the later turns, while most TA-LLM benchmarks, including FunctionChat-Bench, employ teacher-forcing and evaluate each turn independently and thus mitigating error accumulation.

Adding normalization $\psi$ improved slot scores while maintaining relevance scores. As shown in Table \ref{tab:turn_counts}, slot-filling training data consistently features longer chosen trajectory turns compared to rejected ones. Without turn-length normalization $\psi$, this creates artificially inflated reward gaps, potentially hampering effective parameter updates. Conversely, relevance-learning trajectory pairs have longer rejected trajectory turns, explaining the lack of performance gain from the normalization.

The introduction of reward gap margin $\rho$, observed when comparing "DiaTool-DPO w/o \{$\psi$,$\rho$\}", and "DiaTool-DPO w/o $\psi$", resulted in a 4.5\% increase in relevance performance while maintaining slot scores. This aligns with previous research \citep{Meng2024SimPO} showing benefits of reward gap margin subtraction. The different impact on slot versus relevance metrics likely stems from the dominance of aforementioned turn-length bias in slot tasks.

Our complete DiaTool-DPO approach achieved 44\% and 9.6\% improvements over the SFT-only baseline.
The final model reaches 94.8\% of the slot performance of GPT-4o. It also achieves 123.5\% of the relevance score of GPT-4o-mini and 91.3\% of the relevance score of GPT-4o. Notably, call and completion metrics remained comparatively stable throughout ablation studies, confirming that DiaTool-DPO implementation does not compromise tool call capabilities or completion message generation.

\begin{table*}[t]
\footnotesize
\centering
\begin{tabular}{lcccccc}
\toprule
\textbf{Dataset Difficulty} & \textbf{Call} & \textbf{Completion} & \textbf{Slot} & \textbf{relevance} & \textbf{Micro Avg.} & \textbf{Macro Avg.} \\
\midrule
Baseline & 0.843 & \textbf{0.957} & 0.639 & 0.826 & 0.844 & 0.816 \\
\midrule
Easy     & \textbf{0.871} & 0.943          & 0.778          & \textbf{0.913} & 0.850          & 0.876 \\
Hard     & \textbf{0.871} & \textbf{0.957} & 0.778          & \textbf{0.913} & 0.840          & 0.880 \\
All      & 0.857          & 0.929          & \textbf{0.917} & \textbf{0.913} & \textbf{0.905} & \textbf{0.904} \\
\bottomrule
\end{tabular}
\caption{Model performance across different training dataset configurations (Baseline, Easy-only, Hard-only, and All). Evaluation metrics include tool call accuracy, completion rate, slot-filling accuracy, and relevance scores, all reported on a scale of 0 to 1.}
\label{tab:dataset_comparison}
\end{table*}

\subsection{Effect of Dataset Difficulty on Model Performance}
Table \ref{tab:dataset_comparison} compares performance across evaluation metrics for different levels of data difficulty. While DiaTool-DPO trained on both Easy and Hard datasets showed significant improvements over the baseline (SFT-only) model in call, slot, micro average, and macro average metrics, the completion scores remained largely unchanged. This stability in completion scores is reasonable given that this metric primarily evaluates tool call response paraphrasing ability, which is not a primary focus of our dialogue enhancement objectives.

Training separately on Easy and Hard datasets yielded comparable results without clear superiority of either dataset. However, the All dataset, which combines both difficulty levels, demonstrated marked improvements in slot, micro average, and macro average metrics compared to individual training on either Easy or Hard datasets, while maintaining consistent relevance scores. This suggests that slot-filling ability benefits from exposure to a spectrum of difficulty levels rather than exclusively training on either simple or complex examples.

The consistency in relevance scores across Easy, Hard, and All datasets can be attributed to indirect learning transfer: the model's understanding of when to accept tool calls (learned from Type 1 and Type 2 in Easy data) appears to contribute to its ability to identify situations requiring tool call rejection. This finding suggests that the TA-LLM develops an integrated understanding of tool call acceptance and rejection criteria. Nevertheless, the generalizability of this observed phenomenon, wherein training for tool call acceptance inadvertently affects tool call rejection patterns across different TA-LLM architectures, remains to be investigated in Section \ref{sec:different_base_models}. 

\subsection{Effects of SFT and DiaTool-DPO Across Different Base Models}
\label{sec:different_base_models}
\begin{table*}[t]
\centering
\footnotesize  
\begin{tabular}{llcccccc}
\toprule
\textbf{Model} & \textbf{Method} & \textbf{Call} & \textbf{Completion} & \textbf{Slot} & \textbf{relevance} & \textbf{Micro Avg.} & \textbf{Macro Avg.} \\
\midrule
\multirow{3}{*}{Prop.-8B}   & DiaTool-DPO-only   & 0.314          & 0.700          & \textbf{0.833} & 0.609          & 0.575          & 0.614 \\
                                           & SFT-only           & \textbf{0.900} & 0.916          &          0.694 & \textbf{0.913} & 0.870          & 0.856 \\
                                           & SFT + DiaTool-DPO  & 0.886          & \textbf{0.929} & \textbf{0.833} & 0.826          & \textbf{0.884} & \textbf{0.868} \\
\midrule
\multirow{3}{*}{Prop.-3.1B} & DiaTool-DPO-only   & 0.357          & 0.551          & 0.528          &          0.391 & 0.455          & 0.457 \\
                                           & SFT-only           & \textbf{0.771} & 0.817          & 0.750          & \textbf{0.826} & \textbf{0.790} & 0.791 \\
                                           & SFT + DiaTool-DPO  & 0.743          & \textbf{0.871} & \textbf{0.833} & \textbf{0.826} & 0.765          & \textbf{0.818} \\
\midrule
\multirow{3}{*}{LLaMA-3-8B}  & DiaTool-DPO-only   & 0.029          & 0.449          & 0.056          & 0.261          & 0.205          & 0.199 \\
                                      & SFT-only           & 0.843          & \textbf{0.957} & 0.639          & 0.826          & 0.844          & 0.816 \\
                                      & SFT + DiaTool-DPO  & \textbf{0.857} & 0.929          & \textbf{0.917} & \textbf{0.913} & \textbf{0.905} & \textbf{0.904} \\
\bottomrule
\end{tabular}
\caption{Performance comparison across different base models and training methods. Results show the impact of DiaTool-DPO training on Proprietary-8B-Instruct, Proprietary-3.1B-Instruct, and Meta-LLaMA-8B-Instruct models. For each model, we compare DiaTool-DPO-only, SFT-only, and SFT followed by DiaTool-DPO.}
\label{tab:performance_by_base_model}
\end{table*}

While some alignment techniques omit SFT \citep{hong2024orpo}, it typically precedes alignment training in most of the cases \citep{stiennon2020learning,xu2024contrastive,ethayarajh2024kto,ahmadian2024back}. As shown in Table \ref{tab:performance_by_base_model}, we conducted experiments to determine whether SFT is essential before applying DiaTool-DPO.
Since open-source models cannot effectively process Korean without SFT, making evaluation with FunctionChat-Bench challenging, we conducted these experiments on proprietary 8B and 3.1B models trained from scratch 
with Korean language capabilities.
\paragraph{8B-Sized Model.} Without preceding SFT, DiaTool-DPO-only training showed significant performance degradation across all metrics except for slot-filling. This suggests that SFT must precede DiaTool-DPO to establish basic TA-LLM tool-calling capabilities.
The 8B-sized model with SFT + DiaTool-DPO showed a 17\% improvement in slot performance compared to the SFT-only model, but exhibited a 10\% decrease in relevance scores. In our ablation study in Section \ref{subsec:ablation_study}, we observed consistent relevance scores across Easy, Hard, and All datasets, despite Easy data lacking Type 3 data specifically designed for tool call rejection. This led us to question whether indirect learning of tool call rejection from tool call acceptance data was universal across models. The decreased relevance performance of SFT + DiaTool-DPO compared to SFT-only models answers this question, indicating that the ability to indirectly learn tool call rejection is not inherent to all models. As shown in Table \ref{tab:training_data}, Type 3 (tool call rejection) data comprises a mere 14\% compared to Type 1 (tool call) or Type 2 (slot-filling) data. Without indirect learning from tool call acceptance data in Type 1 and Type 2, learning tool call rejection becomes challenging, likely resulting in catastrophic forgetting of relevance capabilities compared to the SFT-only model.
\paragraph{3.1B-Sized Model.} The 3.1B-sized model also demonstrated consistently poor performance across all metrics when trained with DiaTool-DPO without preceding SFT, underscoring the necessity of SFT. However, unlike the 8B-sized model, the 3.1B model showed significant improvements across all performance metrics, including relevance, when comparing SFT-only versus SFT followed by DiaTool-DPO. This suggests that while the 3.1B model may not have fully grasped tool call rejection contexts during SFT, it successfully learned these concepts through Type 3 data in DiaTool-DPO training, demonstrating a different learning pattern from its 8B counterpart.
\subsection{Hyperparameter Analysis}
\label{sec:hyperparameter_analysis}
Effects of hyperparameters $\beta$, $\gamma$ and $\rho$ are analyzed in Appendix~\ref{sec:effect_of_hp}.

\subsection{Extensibility to Other Languages}
\label{sec:paper_qualitative_example}
While this study was primarily conducted in Korean and evaluated using Korean benchmarks, we created an English dataset corresponding to the Easy subset to verify that our approach is applicable to English as well. Although we couldn't perform quantitative evaluation due to the lack of exactly matching English benchmarks, we have included qualitative examples. These examples are appended in Appendix~\ref{sec:qualitative_example}. Also, a detailed analysis in Appendix~\ref{sec:qualitative_example} illustrates how DiaTool-DPO effectively prevents two distinct types of errors commonly observed in SFT-Only baselines.

\section{Conclusion}
In this paper, we introduced DiaTool-DPO, a novel method for enhancing TA-LLMs' conversational capabilities through DPO. Our approach effectively addresses key challenges in dialogue management by formulating the task as a Markov Decision Process with five distinct states and suggesting a noble objective loss that is specified for TA-LLMs. Comprehensive evaluation shows that DiaTool-DPO significantly improves both slot-filling and tool call rejection capabilities while maintaining core functionalities, achieving 94.8\% of GPT-4's slot-filling performance and 91\% of its tool call rejection accuracy. Our analysis demonstrates the benefits of exposure to varied difficulty levels and provides insights into indirect learning transfer between tool call acceptance and tool call reject across different model architectures.

\section*{Limitations}
While we utilized FunctionChat-Bench \citep{kakao2024functionchat} as it appropriately evaluates slot-filling abilities, relevance, and tool call capabilities, its limitation to Korean language prevented quantitative evaluation in other languages. However, we demonstrated qualitative extensibility to English through Section \ref{sec:paper_qualitative_example}.
For future work, we plan to extend our experiments to more suitable benchmarks in other languages as they become available.
To clearly define our problem scope, we did not address scenarios involving multiple simultaneous tool calls such as planning, parallel tool calls, or sequential tool calls. However, our approach can potentially be applied to any task that requires tuning TA-LLMs according to subtle user preferences. Thus, we leave applying this algorithm to more complex problems, including planning and canonical representation \citep{lu2024toolsandbox}, for future work.
For another limitation, the experimental figures in this paper were obtained through a single run.

\section*{Acknowledgements}
I would like to express my gratitude to Kyunghoon Chae, Taesook Jung, Hoon Kim, Daniel Lee, Eunggyun Kim, Myeongcheol Shin, Byungseok Roh, and Dongjin Choi for their invaluable support, which enabled me to pursue my research with joy.

\bibliography{custom}

\appendix
\section{Prompt for Type 2 Trajectory Augmentation from Type 1 Trajectory}
Table \ref{tab:prompt1_easy}, \ref{tab:prompt2_easy} and \ref{tab:prompt3_easy} show GPT prompts for generating Type 2 trajectory by augmenting Type 1 trajectory to create Easy subset of DiaTool-TPO.
Table \ref{tab:prompt1}, \ref{tab:prompt2} and \ref{tab:prompt3} show GPT prompts for generating Type 2 trajectory by augmenting Type 1 trajectory to create Hard subset of DiaTool-TPO.
`gpt-4-turbo-2024-04-09' was used for data augmentation.
\label{sec:slot-fill-prompt}

\begin{table*}[p]
\small
\begin{center}
\begin{tabular}{p{0.95\textwidth}}
\hline
In the first dialogue turn of source sample, the user provides all the information to extract the `required' field of corresponding function.
Thus, in the second dialogue turn of source sample, the assistant had all the argument information to call the `calculate\_tip' function in tool\_calls.
If the user had uttered the utterance in the first turn such that there was missing information about any argument in the `required' fields, the conversation scenario would have been the same as the target sample.
In other words, a couple of more dialogue turns must have been made between the user and assistant to clarify the missing arguments to make the tool call with the `required' arguments filled.
Note that only `required' fields in the function definition needs to be asked. Other arguments are optional and assistant must not ask about optional arguments.
For the following input sample, \\[1ex]

1. Modify the user utterance in the first dialogue turn so that there is not enough information about one or more arguments among the `required' fields.\\
2. Generate following assistant question and user answer utterances accordingly.\\
3. Your answer MUST be in strict JSON format.\\
4. As you can see from the example target sample, generate only ``messages'' part, not, ``tools'' part.\\
5. The generated conversation MUST be in Korean\\
6. Do not change any detail of the conversation if not mentioned otherwise\\[1ex]

\noindent\textbf{\#\#\# Source sample}\\
\begin{minipage}{0.95\textwidth}
\footnotesize
\begin{verbatim}
{
  "messages": [
    {
      "role": "user",
      "content": "Hi, I need to calculate my BMI. I weigh 70 kg and my height is 1.75 m."
    },
    {
      "role": "assistant",
      "content": "Sure, I can help you with that. Let's calculate your BMI.",
      "tool_calls": [
        {
          "id": null,
          "type": "function",
          "function": {
            "name": "calculate_bmi",
            "arguments": "{\"weight\": 70, \"height\": 1.75}"
          }
        }
      ]
    }
\end{verbatim}
\end{minipage}\\
\hline
\end{tabular}
\end{center}
\caption{The prompt for Type 1 to Type 2 data augmentation in DiaTool-DPO Easy subset (Part 1).}
\label{tab:prompt1_easy}
\end{table*}

\begin{table*}[p]
\small
\begin{center}
\begin{tabular}{p{0.95\textwidth}}
\hline
\begin{minipage}{0.95\textwidth}
\footnotesize
\begin{verbatim}
    {
      "role": "tool",
      "content": "{\"bmi\": 22.86}",
      "tool_call_id": null,
      "name": "calculate_bmi"
    },
    {
      "role": "assistant",
      "content": "Your Body Mass Index (BMI) is 22.86. This is considered a healthy weight 
                  for your height."
    }
  ],
  "tools": [
    {
      "type": "function",
      "function": {
        "name": "calculate_bmi",
        "description": "Calculate the Body Mass Index (BMI)",
        "parameters": {
          "type": "object",
          "required": ["weight", "height"],
          "properties": [
            {
              "type": "number",
              "description": "The weight in kilograms"
            },
            {
              "type": "number",
              "description": "The height in meters"
            }
          ]
        }
      }
    }
  ]
}
\end{verbatim}
\end{minipage}\\
\hline
\end{tabular}
\end{center}
\caption{The prompt for Type 1 to Type 2 data augmentation in DiaTool-DPO Easy subset (Part 2).}
\label{tab:prompt2_easy}
\end{table*}

\begin{table*}[p]
\small
\begin{center}
\begin{tabular}{p{0.95\textwidth}}
\hline
\noindent\textbf{\#\#\# Target sample}\\
\begin{minipage}{0.95\textwidth}
\footnotesize
\begin{verbatim}
{
  "messages": [
    {
      "role": "user",
      "content": "Hi, I need to calculate my BMI. My height is 1.75 m."
    },
    {
      "role": "assistant",
      "content": "How much do you weigh?"
    },
    {
      "role": "user",
      "content": "I weigh 70 kg."
    },
    {
      "role": "assistant",
      "content": "Sure, I can help you with that. Let's calculate your BMI.",
      "tool_calls": [
        {
          "id": null,
          "type": "function",
          "function": {
            "name": "calculate_bmi",
            "arguments": "{\"weight\": 70, \"height\": 1.75}"
          }
        }
      ]
    },
    {
      "role": "tool",
      "content": "{\"bmi\": 22.86}",
      "tool_call_id": null,
      "name": "calculate_bmi"
    },
    {
      "role": "assistant",
      "content": "Your Body Mass Index (BMI) is 22.86. This is considered a healthy weight
                  for your height."
    }
  ]
}
\end{verbatim}
\end{minipage}\\[1ex]

\noindent\textbf{\#\#\# Source sample}\\
\hline
\end{tabular}
\end{center}
\caption{The prompt for Type 1 to Type 2 data augmentation in DiaTool-DPO Easy subset (Part 3).}
\label{tab:prompt3_easy}
\end{table*}

\begin{table*}[p]
\small
\begin{center}
\begin{tabular}{p{0.95\textwidth}}
\hline
You are an AI tasked with transforming conversations between a tool-augmented LLM assistant and a user that lack slot-filling conversations (Source) into conversations that include slot-filling conversations (Target).

To understand what slot-filling conversation is, you need to know the following rule for tool-augmented LLMs:
"If the user hasn't mentioned any of the required fields for a corresponding function in the first turn of conversation, the assistant must ask questions to determine all required field values before calling the function. It is prohibited to call the function by arbitrarily filling in required fields or calling the function with empty required fields without discovering all required fields."

A slot-filling question is a question that the assistant asks the user to determine all required fields for a function in a tool-augmented LLM.
The user's answer to this question is called a slot-filling answer.
And together, slot-filling questions and answers are called slot-filling conversation.

When we say slot-filling conversation is unnecessary in the Source, it specifically means that in the first turn of the Source conversation, the user provided all required fields for the function, allowing the assistant to immediately call the appropriate function in the second turn.\\[1ex]

You must generate Target conversations, so please thoroughly understand all the following rules.
Generate your response in strict JSON format.\\[1ex]

\noindent\textbf{Rules}\\
1. In the first conversation turn, the user makes a request to the assistant without including any information about required fields.\\
2. In the second turn, since there are 3 required field arguments but the user's first turn contained 0 arguments, the assistant asks questions about 3-0=3 arguments in the order they appear in the required fields.\\
3. In the third turn, the user randomly answers only 1 out of the 3 questions asked by the assistant.\\
4. In the fourth turn, the assistant asks about the remaining 2 required fields.\\
5. In the fifth turn, the user randomly answers only 1 out of the 2 questions asked by the assistant.\\
6. In the sixth turn, the assistant asks about the remaining 1 question.\\
7. In the seventh turn, the user answers this final question.\\[1ex]

\noindent\textbf{\#\#\#\# Example}\\
\noindent\textbf{\# Source}\\
\begin{minipage}{0.95\textwidth}
\footnotesize
\begin{verbatim}
{
  "messages": [
    {
      "role": "user",
      "content": "Hi, please translate this French sentence into English.
      \"Je suis vraiment heureux de te rencontrer\""
    },
    {
      "role": "assistant",
      "content": "Translation begins.",
      "tool_calls": [
        {
          "id": null,
          "type": "function",
          "function": {
            "name": "translate_text",
            "arguments": "{\"text\": \"Je suis vraiment heureux de te rencontrer\",
            \"source_language\": \"French\",
            \"target_language\": \"English\"}"
          }
        }
      ]
    },
\end{verbatim}
\end{minipage}\\
\hline
\end{tabular}
\end{center}
\caption{The prompt for Type 1 to Type 2 data augmentation in DiaTool-DPO Hard subset (Part 1).}
\label{tab:prompt1}
\end{table*}

\begin{table*}[p]
\small
\begin{center}
\begin{tabular}{p{0.95\textwidth}}
\hline
\begin{minipage}{0.95\textwidth}
\footnotesize
\begin{verbatim}
    {
      "role": "tool",
      "content": "{\"translated_text\": \"I'm really happy to meet you\"}",
      "tool_call_id": null,
      "name": "translate_text"
    },
    {
      "role": "assistant",
      "content": "\"Je suis vraiment heureux de te rencontrer\" translates into 
                  \"I'm really happy to meet you\"
      in English",
      "tool_calls": null
    }
  ],
  "tools": [
    {
      "type": "function",
      "function": {
        "name": "translate_text",
        "description": "Text translation from one language to another.",
        "parameters": {
          "type": "object",
          "required": [
            "text",
            "source_language",
            "target_language"
          ],
          "properties": {
            "text": {
              "type": "string",
              "description": "Text to translate"
            },
            "source_language": {
              "type": "string",
              "description": "Source langauge of the text"
            },
            "target_language": {
              "type": "string",
              "description": "Target language to translate into"
            }
          }
        }
      }
    }
  ]
}
\end{verbatim}
\end{minipage}\\
\hline
\end{tabular}
\end{center}
\caption{The prompt for Type 1 to Type 2 data augmentation in DiaTool-DPO Hard subset (Part 2).}
\label{tab:prompt2}
\end{table*}

\begin{table*}[p]
\small
\begin{center}
\begin{tabular}{p{0.95\textwidth}}
\hline
\noindent\textbf{\# Target}\\
\begin{minipage}{0.95\textwidth}
\footnotesize
\begin{verbatim}
{
  "messages": [
    {
      "role": "user",
      "content": "Hi, I have something to translate."
    },
    {
      "role": "assistant",
      "content": "Please tell me the sentence you want to translate, the source language, 
                  and the target language."
    },
    {
      "role": "user",
      "content": "Translate this into English"
    },
    {
      "role": "assistant",
      "content": "Please tell me the sentence you want to translate and the source language."
    },
    {
      "role": "user",
      "content": "The sentence is \"Je suis vraiment heureux de te rencontrer\"."
    },
    {
      "role": "assistant",
      "content": "Please tell me the source language."
    },
    {
      "role": "user",
      "content": "The source language is French."
    },
    {
      "role": "assistant",
      "content": "Let's start translating.",
      "tool_calls": [
        {
          "id": null,
          "type": "function",
          "function": {
            "name": "translate_text",
            "arguments": "{\"text\": \"Je suis vraiment heureux de te rencontrer\", 
                           \"source_language\": \"French\",
                           \"target_language\": \"English\"}"
          }
        }
      ]
    },
    {
      "role": "tool",
      "content": "{\"translated_text\": \"I'm really happy to meet you\"}",
      "tool_call_id": null,
      "name": "translate_text"
    },
    {
      "role": "assistant",
      "content": "\"Je suis vraiment heureux de te rencontrer\" translates into 
                  \"I'm really happy to meet you\" in English",
      "tool_calls": null
    }
  ]
}
\end{verbatim}
\end{minipage}\\[1ex]

\noindent\textbf{\# Source}\\
\hline
\end{tabular}
\end{center}
\caption{The prompt for Type 1 to Type 2 data augmentation in DiaTool-DPO Hard subset (Part 3).}
\label{tab:prompt3}
\end{table*}

\section{Details of CPT and SFT}
\label{sec:SFT}
During the CPT phase, the language model is trained using next-token prediction loss on an open-source tool call dataset. 
During SFT, user turns are masked, and the loss is backpropagated only through the assistant turns' next-token predictions.

To leverage widely available open-source data, we utilize English datasets for CPT, while the SFT phase uses Korean-translated datasets. 
The details of dataset usage for each phase are available in Table \ref{tab:cpt_sft_training_data}.
`schema\_guided\_dstc8' refers to Schema-Guided Dialogue dataset which include interactions with services and APIs spanning 20 domains, such as banking, events, media, calendar, travel, and weather \citep{rastogi2019towards}. `xlam-function-calling-60k' refers to the work of \citet{liu2024apigen} which generates this dataset through automated data generation pipline named APIGen.
\begin{table}[h]
\centering
\scriptsize
\begin{tabular}{llcc}
\toprule
Stage & Dataset Name & Train Set Size & Language \\
\midrule
\multirow{3}{*}{CPT} & glaive2.0 & 110K & English \\
& xlam-function-calling-60k & 60K & English \\
& schema\_guided\_dstc8 & 16K & English \\
\midrule
\multirow{2}{*}{SFT} & glaive2.0 & 47K & Korean \\
& schema\_guided\_dstc8 & 16K & Korean \\
\bottomrule
\end{tabular}
\caption{Training datasets used in each stage. English datasets are employed for CPT while Korean datasets are used for SFT. Dataset sizes are reported in thousands (K) of examples.}
\label{tab:cpt_sft_training_data}
\end{table}
\section{Training Setup}
\label{sec:training_setup}
All experiments were conducted on 8 NVIDIA A100-80GB GPUs, with training completed in approximately 2.5 hours. For training, we used the AdamW optimizer $(\beta_1 = 0.9, \beta_2 = 0.999, \epsilon = 1e-8,$ weight decay = 0.0) with a learning rate of 1e-7 and mixed-precision training with bfloat16 to optimize memory usage and computational efficiency. We employed a total batch size of 8 distributed across the GPUs. The sequences were configured with a maximum total length of 8,192 tokens, allowing prompts up to 4,096 tokens. The model was trained for a single epoch using a linear learning rate scheduler with 150 warmup steps. Unless otherwise specified, we maintained consistent hyperparameters across experiments with $\beta = 0.5,$  $\gamma = 0.5,$ and $\rho = 2.0$. All other hyperparameters followed the default settings of the DPOConfig in the TRL library \citep{vonwerra2022trl}. Also, for the tool conversation template, we used functionary's v3.llama3 \citep{meetkai2024functionary}.

We created the DiaTool-DPO dataset based on the glaive-2.0 dataset \citep{glaive2023function}, which is licensed under Apache 2.0. Additionally, we developed the DiaTool-DPO source code based on open-source TRL \citep{vonwerra2022trl} and Functionary 
 \citep{meetkai2024functionary}, which are licensed under Apache 2.0 and MIT licenses respectively. Our research has properly utilized these licenses within their permitted scope.
\section{Hyperparameter Analysis}
\label{sec:effect_of_hp}
\begin{figure*}[t]
    \centering
    \includegraphics[width=\textwidth]{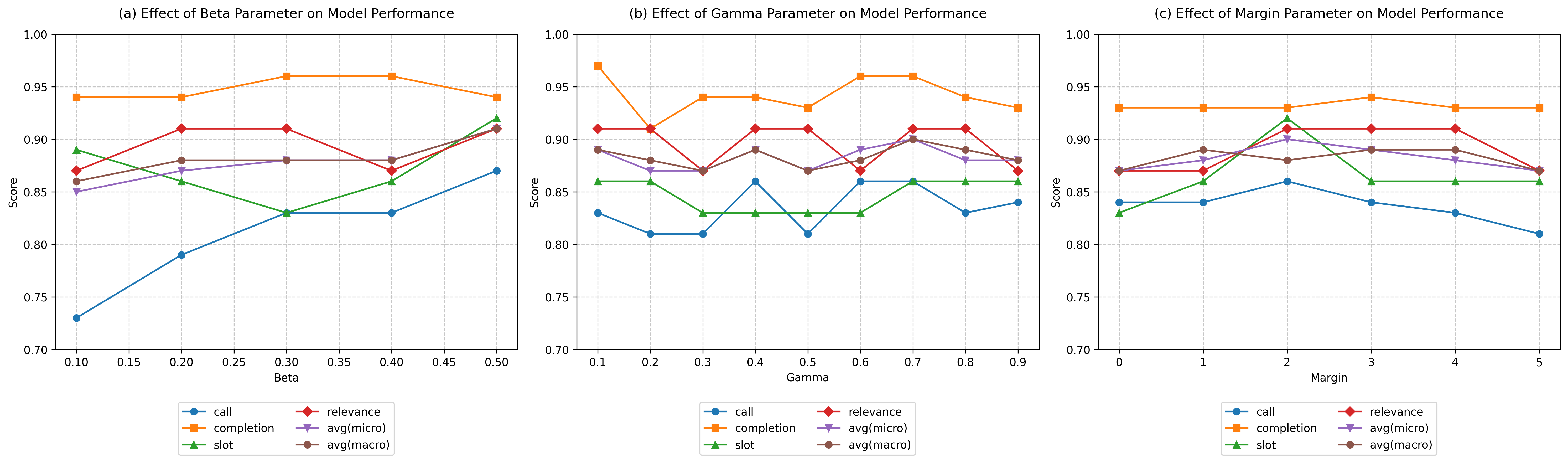}
    \caption{Effects of hyperparameters on model performance metrics. (a) Impact of DPO regularization parameter ($\beta$) ranging from 0.1 to 0.5. (b) Impact of reward scaling factor ($\gamma$) from 0.1 to 0.9. (c) Impact of reward gap margin ($\rho$) from 0 to 5. All experiments measure six different performance metrics: call accuracy, completion accuracy, slot accuracy, relevance accuracy, and micro/macro-averaged scores.}
    \label{subplot_horizontal}
\end{figure*}
\paragraph{Impact of $\beta$ Value}
As depicted in Figure \ref{subplot_horizontal} (a), experimental results demonstrate that increasing the $\beta$ parameter leads to consistent improvements in call performance, with corresponding gains in both micro and macro averages. Conversely, models trained with lower $\beta$ values exhibit overfitting to slot-filling and relevance tasks while showing catastrophic forgetting of tool call capabilities previously acquired during supervised fine-tuning.
\paragraph{Impact of $\gamma$ Value}
Variations in the $\gamma$ parameter did not yield statistically significant differences in performance metrics. As previously discussed in Section \ref{subsec:ablation_study}, we hypothesize that this phenomenon occurs because the evaluation methodology of FunctionChat-Bench is based on teacher-forcing, which prevents error propagation from previous turns, thereby making it insensitive to variations in the $\gamma$ value.
\paragraph{Impact of $\rho$ Value}
We observed a negative correlation between margin size and call performance. Since margin serves as a reward gap threshold, larger values intensify the DiaTool-DPO training signal. This phenomenon mirrors the behavior observed with small $\beta$ values, suggesting that excessive margins lead to catastrophic forgetting of fundamental tool call capabilities acquired during supervised fine-tuning.


\section{Extensibility to Other Languages}
\begin{figure*}[t]
    \centering
    \includegraphics[width=\textwidth]{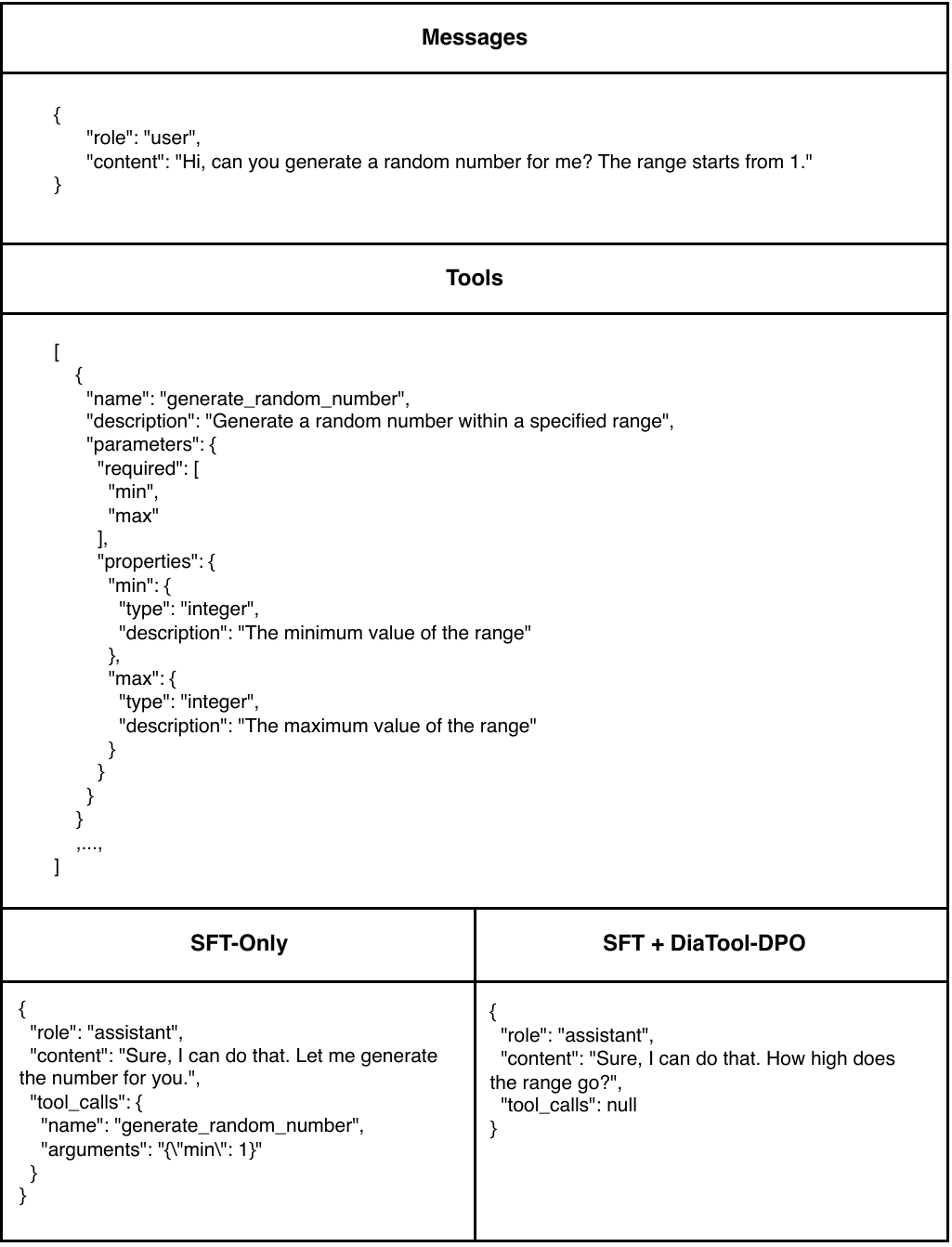}
    \caption{Comparison of responses between SFT-Only and SFT + DiaTool-DPO models. The `Messages' shows the user's initial query, and `Tools' presents the tool specification required to resolve the user's query. For brevity, we omit the remaining 4-6 candidate tools in Tools.    
    While the tool specification includes `min' and `max' as required fields, the user only specifies the `min' value in the query. The SFT-Only model proceeds to call the tool using only the min value, whereas SFT + DiaTool-DPO model engages in slot-filling by asking for the missing `max' value.}
    \label{fig:qualitative_example}
\end{figure*}

\begin{figure*}[t]
    \centering
    \includegraphics[width=\textwidth]{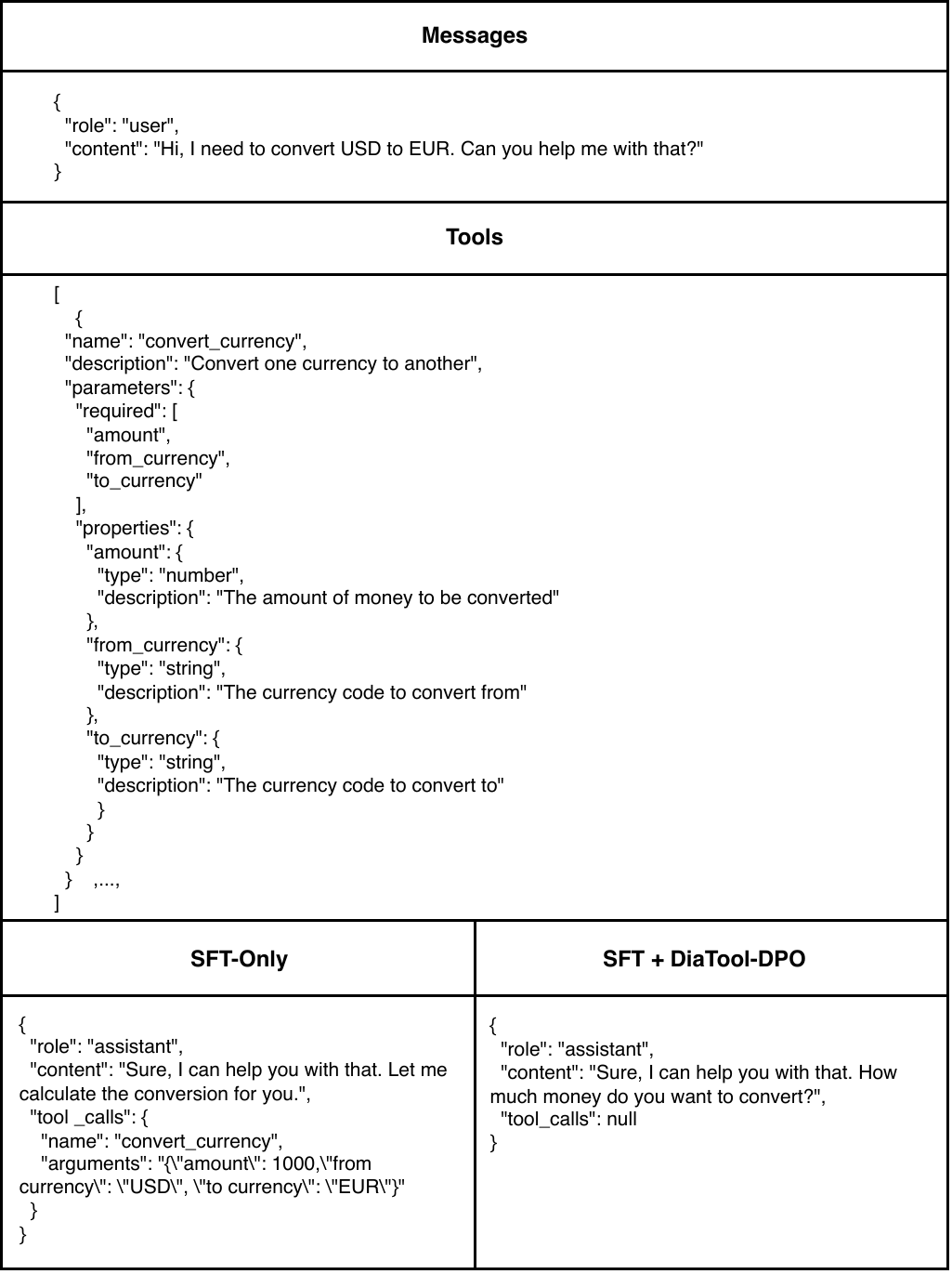}
    \caption{Comparison of responses between SFT-Only and SFT + DiaTool-DPO models. The `Messages' shows the user's initial query, and `Tools' presents the tool specification required to resolve the user's query. For brevity, we omit the remaining 4-6 candidate tools in Tools.
    While the tool specification lists `amount', `from\_currency', and `to\_currency' as required fields, the user's query lacks the `amount' information. The SFT-Only model hallucinates a value of 1,000 for the amount field, whereas the SFT + DiaTool-DPO model engages in slot-filling by asking a question to determine the missing amount value.}
    \label{fig:qualitative_example2}
\end{figure*}

Our approach is not limited to Korean. We qualitatively validate its extensibility to English, with examples shown in Figures \ref{fig:qualitative_example} and  \ref{fig:qualitative_example2}.
Through qualitative analysis, we demonstrate distinct error patterns in how the SFT-Only model processes missing required fields. In Figure \ref{fig:qualitative_example}, we observe that the model fails to recognize a missing required field and simply omits it without generating a slot-filling question. A more concerning behavior is illustrated in Figure \ref{fig:qualitative_example2}, where the model hallucinates the missing required field `amount' with an arbitrary value of 1,000. This case is particularly concerning as incorrect tool calls can be made without user awareness.

\label{sec:qualitative_example}

\end{document}